\documentclass[sigconf]{acmart}

\copyrightyear{2023}
\acmYear{2023}
\setcopyright{acmlicensed}
\acmConference[MM '23] {Proceedings of the 31st ACM International Conference on Multimedia}{October 29--November 3, 2023}{Ottawa, ON, Canada.}
\acmBooktitle{Proceedings of the 31st ACM International Conference on Multimedia (MM '23), October 29--November 3, 2023, Ottawa, ON, Canada}
\acmPrice{15.00}
\acmISBN{979-8-4007-0108-5/23/10}
\acmDOI{10.1145/3581783.3611982}

\usepackage{soul}
\usepackage{url}
\usepackage{graphicx}
\usepackage{amsmath}
\usepackage{amsthm}
\usepackage{booktabs}
\usepackage{algorithm}
\usepackage{newtxmath}
\usepackage{multirow}
\usepackage{setspace}
\usepackage{color}
\usepackage{balance}
\newcommand{\etal}{\textit{et al}.}
\newcommand{\ie}{\textit{i}.\textit{e}.}
\newcommand{\eg}{\textit{e}.\textit{g}.}
\newcommand{\etc}{\textit{etc}}

\AtBeginDocument{%
  }

\acmSubmissionID{1293}

\begin{document}
\begin{sloppypar}
\title{Point-aware Interaction and CNN-induced Refinement Network for RGB-D Salient Object Detection}

\author{Runmin Cong}

\authornote{Runmin Cong, Wei Zhang and Ran Song are also affiliated with the Key Laboratory of Machine Intelligence and System Control, Ministry of Education, Jinan, Shandong, China.}
\orcid{0000-0003-0972-4008}
\affiliation{%
  \institution{Beijing Jiaotong University \& Shandong University}
  \city{Beijing \& Jinan}
  \country{China}
}
  \email{rmcong@sdu.edu.cn}

\author{Hongyu Liu}
\authornote{Hongyu Liu and Chen Zhang are  affiliated with Institute of Information Science, Beijing Jiaotong University and the Beijing Key Laboratory of Advanced Information Science and Network Technology.}
\orcid{0000-0001-8038-7636}

\author{Chen Zhang}
\authornotemark[2]
\authornote{Corresponding author.}
\orcid{0000-0002-0492-2489}

\affiliation{%
  \institution{Beijing Jiaotong University}
  \city{Beijing}
  \country{China}
  \postcode{100044}
}
\email{{liu.hongyu,chen.zhang}@bjtu.edu.cn}


\author{Wei Zhang}
\orcid{0000-0002-4960-3190}
\authornotemark[1]
\affiliation{%
  \institution{Shandong University}
  \city{Jinan}
  \state{Shandong}
  \country{China}
  \postcode{250061}
}
\email{davidzhang@sdu.edu.cn}

\author{Feng Zheng}
\orcid{0000-0002-1701-9141}

\affiliation{%
  \institution{Southern University of Science and Technology}
  \city{Shenzhen}
  \state{Guangdong}
  \country{China}
  \postcode{518055}
}
\email{f.zheng@ieee.org}

\author{Ran Song}
\orcid{0000-0002-1344-4415}
\authornotemark[1]

\affiliation{%
  \institution{Shandong University}
  \city{Jinan}
  \state{Shandong}
  \country{China}
  \postcode{250061}
}
\email{ransong@sdu.edu.cn}

\author{Sam Kwong}
\orcid{0000-0001-7484-7261}

\affiliation{%
  \institution{City University of Hong Kong}
  \city{Hong Kong SAR}
  \country{China}
  \postcode{51800}
}
\email{cssamk@cityu.edu.hk}

\renewcommand{\shortauthors}{Runmin Cong, et al.}

\begin{abstract}
 By integrating complementary information from RGB image and depth map, the ability of salient object detection (SOD) for complex and challenging scenes can be improved. In recent years, the important role of Convolutional Neural Networks (CNNs) in feature extraction and cross-modality interaction has been fully explored, but it is still insufficient in modeling global long-range dependencies of self-modality and cross-modality. To this end, we introduce CNNs-assisted Transformer architecture and propose a novel RGB-D SOD network with Point-aware Interaction and CNN-induced Refinement (PICR-Net). On the one hand, considering the prior correlation between RGB modality and depth modality, an attention-triggered cross-modality point-aware interaction (CmPI) module is designed to explore the feature interaction of different modalities with positional constraints. On the other hand, in order to alleviate the block effect and detail destruction problems brought by the Transformer naturally, we design a CNN-induced refinement (CNNR) unit for content refinement and supplementation. Extensive experiments on five RGB-D SOD datasets show that the proposed network achieves competitive results in both quantitative and qualitative comparisons. Our code is publicly available at:  \textit{\url{https://github.com/rmcong/PICR-Net_ACMMM23}.}
\end{abstract}




\maketitle

\section{Introduction}
Inspired by the human visual system, Salient Object Detection (SOD) aims to locate the most attractive objects or regions in a given scene \cite{crm/tcsvt19/review,surveyfan,crm/tcsvt22/weaklySOD,crm/aaai20/GCPANet,crm/tip21/DAFNet,crm/acmmm21/light-field,crm/tnnls22/360SOD,crm/tgrs22/RRNet,crm/tmm22/TNet,crm/tcyb22/glnet,crm/tetci22/PSNet}, which has been successfully applied to abundant tasks. Furthermore, the RGB-D SOD task additionally introduces the depth map in the SOD task to better simulate the ability of the human binocular vision system and obtain the ability to perceive the distance relation between objects.  
Since entering the era of deep learning \cite{crm/tip20/MCMT-GAN,crm/spl21/underwater,crm/access17/dsr, zhang2023controlvideo}, the convolutional neural networks (CNNs) based RGB-D SOD frameworks have been vigorously developed \cite{A2dele,BBS-Net,UC-Net,DCF,crm/tcyb21/ASIFNet,crm/acmmm21/CDINet,MVSalNet,crm/tip21/DPANet,crm/tip21/DynamicRGBDSOD,cmMS,crm/tip22/CIRNet}, far surpassing the performance of hand-crafted feature based methods.
However, it is shown that although CNNs can theoretically obtain a larger receptive field through network deepening, in practice, the receptive field of convolutional operation is still limited to a local scope. 
And unfortunately, the determination of salient objects requires global contrastive perception, so the ability to model global relation plays a crucial role in salient object detection.
Therefore, some research works leverage the global modeling capability of Transformer to realize RGB-D SOD~\cite{TriTransNet,vst}.


\begin{figure}[!t]
  \centering
  \includegraphics[width=\linewidth]{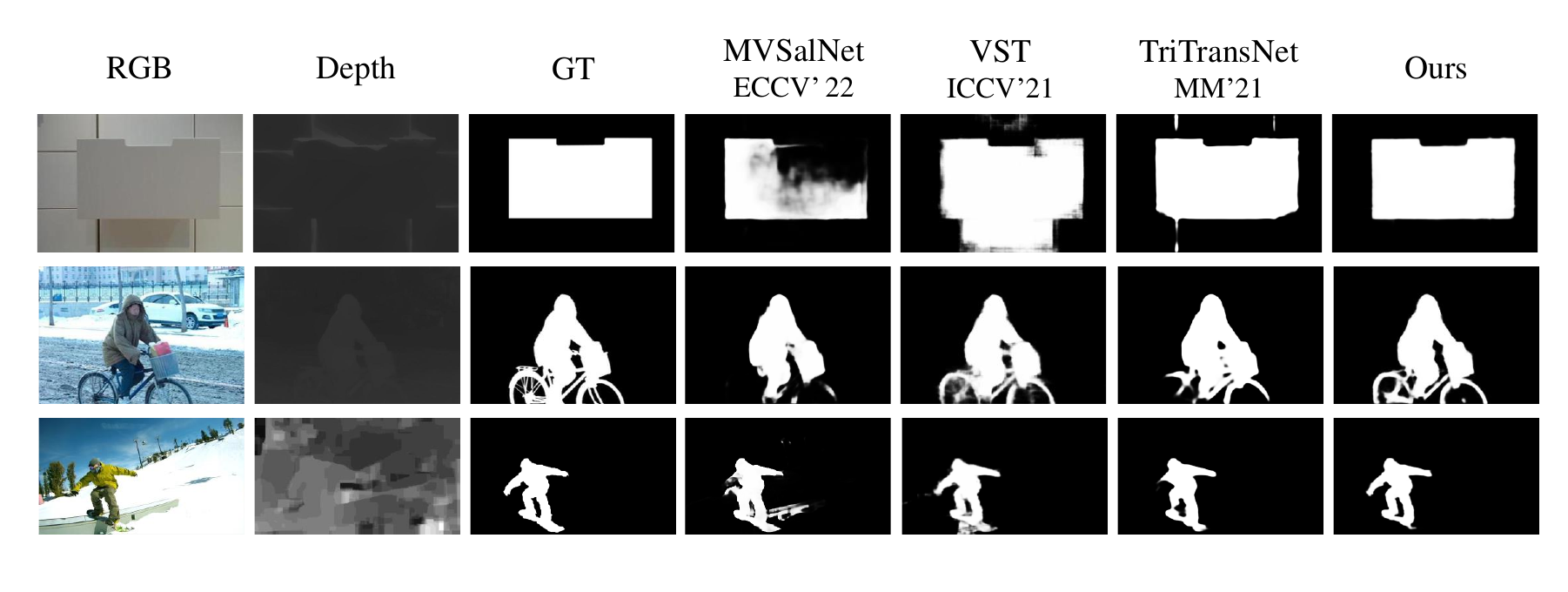}
  \caption{Visual comparison of representative networks with different  architectures,  where  MVSalNet~\cite{MVSalNet}, VST~\cite{vst} and TriTransNet~\cite{TriTransNet} are the pure CNNs, pure Transformer, and Transformer-assisted CNNs architectures, respectively.}
  \label{trans_cnn}
\end{figure}

From the perspective of model architecture, the existing RGB-D SOD methods can be divided into three categories: pure CNNs model, pure Transformer model, and Transformer-assisted CNNs model. 
For the pure CNNs architecture, thanks to the excellent local perception ability of the convolutional operation, the saliency results perform better in describing some local details (such as boundaries), but may be incomplete, such as the result of MVSalNet \cite{MVSalNet} in the first image of Figure~\ref{trans_cnn}.
For the pure Transformer structure, since the Transformer can capture long-range dependencies, the integrity of the detection results is improved to a certain extent, but the patch-dividing operation may destroy the quality of details, induce block effects, and even introduce additional false detections, such as the result of VST \cite{vst} in Figure~\ref{trans_cnn}.
The Transformer-assisted CNNs structure introduces the Transformer to assist CNNs for global context modeling, which can alleviate the shortcomings of the above single scheme by combining the two. However, in the process of decoding layer by layer, the convolution operation will gradually dilute the global information obtained by Transformer, so this scheme will still lead to the missing or false detection, such as the result of TriTransNet \cite{TriTransNet} in Figure~\ref{trans_cnn}.  
Therefore, in this paper, we rethink the relationship between Transformer and CNNs and propose a CNNs-assisted Transformer network architecture. Specifically, we utilize Transformer to complete most of the encoding and decoding process, and design a pluggable CNN-induced refinement (CNNR) unit to achieve content refinement at the end of the network. 
In this way, the Transformer and CNNs can be fully utilized without interfering with each other, thereby gaining global and detail perception capabilities and generating accurate and high-quality saliency map.

For the cross-modality feature interaction issue, traditional feature interaction mechanism has raised great attention from computer vision and pattern recognition, even with successful achievements when the modality correspondance message is missing~\cite{jiang2019dm2c, wen2021seeking}. Under the context of Transformer-based models, the cross-attention scheme~\cite{CrossAttention,cit2.12071} is a commonly used method.
For example, 
the cross-modality interaction in vision-language task calculates similarities between different modalities by alternating queries and keys from vision and language modalities.
Likewise, the cross-attention mechanism can also be directly applied to the RGB-D SOD task to model the relation between RGB and depth features, but there are two main challenges.
 First, unlike the relation between image and language, RGB image and depth map only have clear correlations in the features of the corresponding positions, so the above cross-attention approach is somewhat blind and redundant. 
 Second, since the computational complexity is quadratically proportional to the size of the feature map, this undifferentiated all-in-one calculation will bring unnecessary computational burden.
To address the above two issues, we propose a cross-modality point-aware interaction (CmPI) module, which simplifies the modeling process of cross-modality interactions by grouping corresponding point features from different modalities. 
In this way, the interaction of RGB and depth features is constrained to the same position, making it more directional and reducing the computational complexity to a linear level.
In addition, we also introduce global saliency guidance vectors in CmPI to emphasize global constraint while conducting cross-modality interaction, making the interaction more comprehensive. Specifically, a two-step attention operation with well-designed mask constraints is used to achieve above cross-modality and global-local relation modeling process.   


In general, our paper makes three major contributions:

\begin{itemize}
\item To take full advantage of both Transformer and CNNs, we propose a new CNNs-assisted Transformer architecture to achieve RGB-D SOD, denoted as PICR-Net, which wins competitive performance against 16 the state-of-the-art methods on five widely-used datasets.
  \item Considering the priori correlation between RGB modality and depth modality, we propose a cross-modality point-aware interaction module that dynamically fuses the feature representations of different modalities under global guidance and location constraint.
  \item To alleviate the block effect and detail destruction problems caused by the Transformer architecture, we design a pluggable CNN-induced refinement unit at the end of the network to achieve content refinement and detail supplement.
\end{itemize} 

\begin{figure*}[!t]
\centering
\centerline{\includegraphics[width=0.94\textwidth]{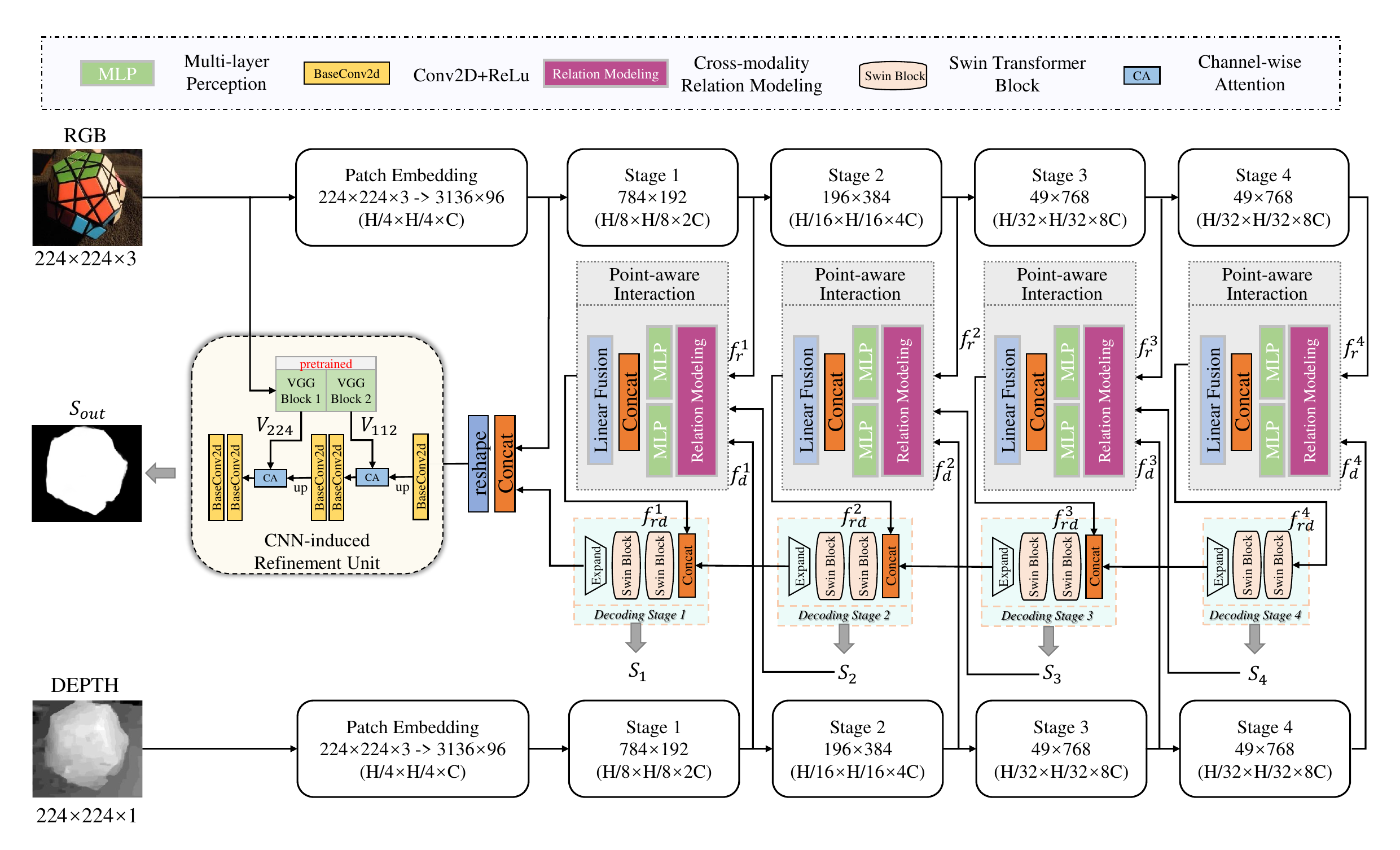}}
\caption{The overall framework of the proposed PICR-Net. First, RGB image and depth image are fed to a dual-stream encoder to extract corresponding multi-level features $\big\{f_{r}^{i}\big\}_{i=1}^{4}$ and $\big\{f_{d}^{i}\big\}_{i=1}^{4}$. 
Subsequently, the features of the same layer are multi-dimensionally interacted through cross-modality point-aware interaction module, where the previously output saliency map $S_{i+1}$ is used to extract global guidance information. At the end of the network, the CNNR unit provides convolutional features with higher resolution and more detail from the pre-trained VGG16 model to refine and output the final high-quality  saliency map $S_{out}$.}
\label{fig_overall}
\end{figure*}


\section{Related Work}

In early days, the traditional RGB-D SOD methods~\cite{tradition1,tradition2,crm/tcyb20/going} rely on hand-crafted features and have very limited performance.
In recent years, thanks to the powerful feature representation ability of deep learning, a large number of learning-based RGB-D SOD models have been proposed.
Before the launch of Vision Transformer in 2020 \cite{vit}, the RGB-D SOD task still uses CNNs as the mainstream architecture, and various models \cite{D3Net,ATSA,BTSNet,SPNet,crm/tip21/DPANet,crm/acmmm21/CDINet,RD3D,Lightweight,JLDCF} have been proposed in terms of cross-modality interaction, depth quality perception, and lightweight design, \etc.
For example, Zhang \etal ~\cite{crm/acmmm21/CDINet} designed a cross-modality discrepant interaction strategy to achieve efficient integration in the RGB-D SOD task. 
Cong \etal ~\cite{crm/tip21/DPANet} considered the quality of depth map in the RGB-D SOD task, and proposed a depth potentiality-aware gated attention network to address the negative influence of the low-quality depth map.
Chen \etal ~\cite{RD3D} stacked 3-D convolutional layers as encoder to achieve RGB-D SOD, which can fuse the cross-modality features effectively without dedicated or sophisticated module. 
Huang \etal ~\cite{Lightweight} performed cross-modal feature fusion only
on one certain level of features rather than on all of the
levels to form a lightweight model. 

As the Transformer shines in the computer vision field, some pure Transformer or a combination of Transformer and CNNs have emerged. 
Liu \etal ~\cite{vst} designed a pure Transformer architecture for RGB-D SOD task from a new perspective of sequence-to-sequence modeling, in which the cross-attention are used for cross-modality interaction. Song \etal ~\cite{MMT} fully used self-attention and cross-attention for interaction between appearance features and geometric features in the RGB-D SOD task. 
Liu \etal ~\cite{TriTransNet} embedded the Transformer after the CNNs to model the long-range dependencies between convolutional features and achieve fusion at the same time.

However, these existing pure CNNs or pure Transformer solutions also have some problems. For example, the CNNs-based methods are somewhat inferior in the ability to acquire global information to accurately locate salient objects, while the Transformer-based solutions are computationally intensive and susceptible to block effects.
Although some methods using hybrid structure following Transformer-assisted CNNs architecture can alleviate the above concerns to some extent, the multi-layer convolutions during decoding can dilute the global information acquired by Transformers and affect the prediction performance.
We should reconsider the role of Transformers and CNNs in the network, make full use of their respective advantages, and explore effective ways of cross-modality interaction. 
So we try to use a CNNs-assisted Transformer architecture to model global context and local details, and propose a point-aware interaction mechanism under location constraints to make cross-modality interaction more efficient and targeted.

\section{Proposed Method}

\subsection{Network Overview}

As shown in Figure \ref{fig_overall}, the proposed network follows an encoder-decoder structure as a whole. The top and bottom branches are the feature encoders for RGB image and depth map respectively, both of which adopt the shared-weight Swin-Transformer model~\cite{swin}, while the middle branch is the bottom-up decoding process. 
In each decoding stage, the cross-modality representation is firstly obtained by modeling the interaction relation at the same location of different modalities through the CmPI module. 
Thereafter, we use the Swin-Transformer-based decoding blocks to model the long-range dependencies of cross-modality features during decoding from a global perspective. 
Specifically, the cross-modality features $f_{r d}^{i}$ generated by the CmPI module and the upsampled output features $f_{ {decoder}}^{i+1\uparrow}$ of the previous decoding stage (if any) are fed into two cascaded Swin-Transformer blocks to model the global relation: 
\begin{equation}
\begin{split}
\small
      &f_{{decoder }}^{i}=\\
      &\left\{\begin{array}{l}
{Exp}\left(ST\left(f_{r d}^{i}\right)\right), i=4 \\
{Exp}\left({ST}\left({Linear}\left({cat}\left(f_{r d}^{i}, f_{ {decoder }}^{i+1\uparrow}\right)\right)\right)\right), i=\{1,2,3\}
\end{array}\right.,  
\label{dec}
\end{split}
\end{equation}
where $cat$ means the concatenation operation in the feature dimension, $Linear$ is the linear layer, $ST$ represents two Swin-Transformer blocks, and $Exp$ is the operation that converts features back to spatial resolution.
Finally, at the end of the decoder, a pluggable CNNR unit is proposed to address the problems of block effect and detail destruction under the Transformer architecture at a low cost, and generate the final saliency map $S_{out}$.

\begin{figure*}[!t]
\centering
\centerline{\includegraphics[width=0.95\textwidth]{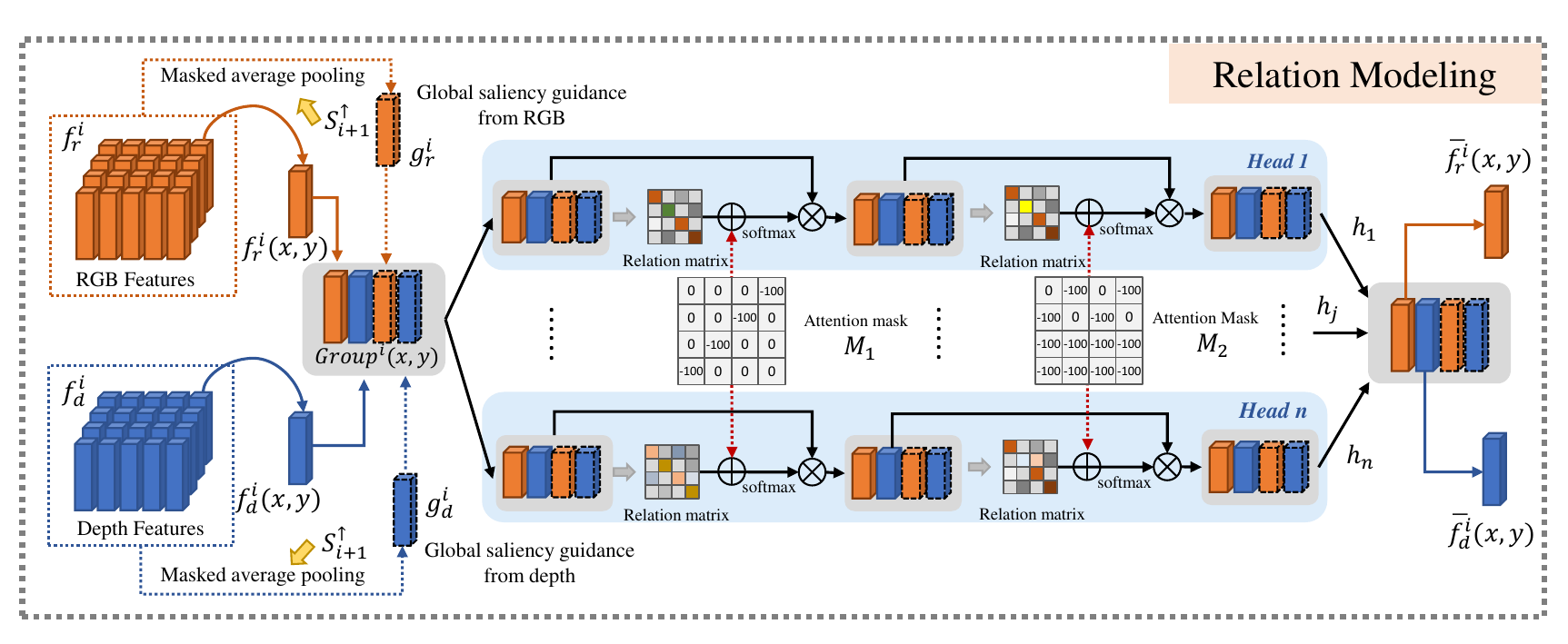}}
\caption{The cross-modality point-aware RM in the CmPI module, where the RGB and depth  features at the same spatial location and the global saliency guidance vectors from both modalities are interacted sufficiently and efficiently.}
\label{fig4}
\end{figure*}

\subsection{Cross-modality Point-aware Interaction Module}
After extracting the multi-level encoding features of RGB modality and depth modality, how to achieve comprehensive interaction is an important issue that needs to be focused on in the encoding stage. 
The existing cross-modality interaction scheme under the Transformer architecture usually models the relation among all positions of two modalities.
But as we all know, there is a corresponding relation between the RGB image and depth map itself, that is, the two modalities have a clear relation only at the corresponding position. 
As such, there is computational redundancy if the relation between all pixels of different modalities is modeled, and unnecessary noise may also be introduced due to this forced association modeling.
Considering these, proceeding from reality of cross-modality modeling in RGB-D SOD task, we introduce the position constraint factors and propose a cross-modality point-aware interaction scheme, the core of which is to explore the interaction relation of different modality features at the same location through the multi-head attention. 
Compared with the direct combination of feature vectors, the multi-head parallel attention allows dynamic interaction of cross-modality features in different embedding spaces, enabling adaptive adjustment of the involvement of two modality features in different scenes.
Moreover, in order to guide this interaction process from a global perspective and perceive the role of the current location in the overall feature map, we also add global saliency guidance vectors to the interaction process.

Figure \ref{fig4} depicts the most critical cross-modality point-aware Relation Modeling (RM) in the CmPI module. Let the point feature vectors corresponding to any location $(x, y)$ on the features of the RGB modality and depth modality be denoted as $ f_{r}^{i}(x, y) \in \mathbb{R}^{1 \times c}$ and $ f_{d}^{i}(x, y) \in \mathbb{R}^{1 \times c}$, where $c$ is the embedding dimension. First, in order to provide a global guidance for the interaction process at each location, the saliency guidance vectors of two modalities are generated by using the upsampled side-output saliency map $S_{i+1}^{\uparrow}$ decoded from the previous level, shared by all locations at the current scale in the computational process:
\begin{equation}
       g_{r}^{i}=M A P\left(f_{r}^{i}, S_{i+1}^{\uparrow}\right), g_{d}^{i}=M A P\left(f_{d}^{i}, S_{i+1}^{\uparrow}\right),
\end{equation}
where $MAP$ represents the masked average pooling \cite{MAP}, and $S_{i+1}^{\uparrow}$ is used as a weighting mask. 
Then, the RGB/depth features at location $(x, y)$ and the RGB/depth saliency guidance vectors together form a point-wise feature group ${Group}^i(x, y) \in \mathbb{R}^{4 \times d}$ with more comprehensive representation:
\begin{equation}
       {Group}^i(x, y)={Stack}\left(f_{r}^{i}(x, y), f_{d}^{i}(x, y), g_{r}^{i}, g_{d}^{i}\right),
\end{equation}
where $Stack$ means to stitch features together into a new dimension. 

Afterwards, the interaction between point feature groups is performed by a relation modeling operation:
\begin{equation}
       \{\bar{f}_{r}^{i}(x, y), \bar{f}_{d}^{i}(x, y)\}=R M_{(x, y)}\left({Group}^i(x, y)\right)[:2],
\end{equation}
where $RM_{(x, y)}$ is the relation modeling operation between RGB and depth modalities at position $(x, y)$, which can be defined as: 
\begin{equation}
       R M_{(x, y)}\left(f_{r}^{i}, f_{d}^{i}\right)= {Linear}\left({ cat }\left( { h }_{1}, \ldots,  { h }_{n}\right) \right),
\end{equation}
where $\big\{h_{j}\big\}_{j=1}^{n}$ represent the attention output results of different heads, \ie, different feature spaces. 
The relation modeling operation is similar to the multi-head attention mechanism \cite{Attention}, but there are also obvious differences:
On the one hand, not all features within the feature group need to be interacted, such as between the guidance vector and features of different modalities (\ie, $f_{r}^{i}(x, y)$ and $g_{d} ^{i}$, $f_{d}^{i}(x, y)$ and $g_{r}^{i}$). Because they are in different scales and from different modalities, forcing interactions can have negative effects instead.
Therefore, we introduce a carefully designed mask in attention operation to suppress such negative interactions. 
On the other hand, after the attention interactions within the feature group, the global vector is updated by other cross-modality global vectors as well as self-modality local vectors. To make better use of this information and emphasize the role of global-local guidance, we also perform a second-step global-local interaction in the self-modality using a new mask constraint.
The above process can be expressed by the following formula:
\begin{equation}
       { h }_{j}={Attention}\left({Attention}\left({Group}_j^i(x, y), M_1\right),M_2\right).
\end{equation}
In the first-step attention calculation, $M_1$ is set to an anti-angle matrix with the value of -100.0, which allows the negative effects of the depth guidance vector on the RGB features and the RGB guidance vector on the depth features to be weaken during the interaction. Afterwards, the second-step attention operation follows, in which the global-local interaction within the self-modality is performed by setting $M_2$ as the value in Figure ~\ref{fig4}, thereby strengthening the guidance of the global vectors on the local representation in the same modality. 
Specifically, the attention operation with the mask is performed as follows:
\begin{equation}
       {Attention}\left({Group}_j^i(x, y), M\right)={softmax}\left(\frac{Q_{j} K_{j}^{T}}{\sqrt{d}}+M\right) V_{j},
\end{equation}
where $Q_{j}$, $K_{j}$ and $V_{j}$ are all generated by linear mapping from ${Group}_j^i(x, y)$, and $j$ is the index of the attention head. 

After the above process, the information of the two modalities can be fully interacted under the guidance of the saliency guidance vector, and finally the two features are combined by a linear layer as the final cross-modality features:
\begin{equation}
       f_{r d}^{i}(x, y)={Linear}\left({cat}\left(M L P\left(\bar{f}_{r}^{i}(x, y)\right), M L P\left(\bar{f}_{d}^{i}(x, y)\right)\right)\right),
\end{equation}
where $MLP$ is the multi-layer perceptron.

\subsection{CNN-induced Refinement Unit}

At the output of the Transformer decoder, the main body of the salient object is basically determined, but due to the patch-dividing in the Transformer structure, the obtained saliency map may have problems of block effect and detail destruction. To this end, we propose a pluggable CNN-induced refinement unit at the end of the decoder. This is mainly inspired by the advantages of the CNNs in processing local details. Moreover, the feature resolution at this stage is larger, and the convolution operation is more reasonable in terms of the number of parameters and computational cost.
Because the main purpose of this step is detail content refinement, there is no need to introduce the complete encoder-decoder network of CNNs, only the shallow features of the first two layers with rich texture details in VGG16 \cite{vgg} are enough, denoted as $V_{224} $ and $V_{112}$.
First, the decoder features $f_{ {decoder}}^{1}$ from the last Transformer layer are converted to pixel level and upsampled to same resolution as $V_{112}$ in preparation for the following refinement:
\begin{equation}
       T_{112}=up\left( { Baseconv }\left( {Exp}\left(f_{ {decoder }}^{1}\right)\right)\right),
\end{equation}
where $Baseconv$ consists of a $3\times3$ convolution layer as well as a ReLu activation function following, and $up$ represents upsampling operation. Thereafter, $V_{224}$ and $V_{112}$ are used for further recovery of resolution.  Considering that simply using concatenation to fuse features can not effectively capture the detail information embedded in certain channels, we use the channel attention \cite{ca} to discover those important channels with detail information while preserving the main body of salient objects for adaptive fusion. 
The progressive refinement process can be expressed as follows:
\begin{equation}
       T_{224}  = { up }\left( { Baseconv }(CA\left({cat}\left(T_{112}, V_{112}\right)\right)\right)) ,
\end{equation}
\begin{equation}
       S_{out}  ={Baseconv}(CA\left({cat}\left(T_{224}, V_{224}\right)\right)),
\end{equation}
where $CA$ denotes the channel attention operation with residual connection, and $S_{out}$ is the final saliency map. In the way, the fine-grained information from convolution can be supplemented for more accurate saliency map.

\subsection{Loss Function}
In order to obtain high quality saliency map with clear boundaries, the proposed entire network is supervised by a mixture of losses, including the commonly used binary cross-entropy loss, the SSIM loss for measuring structural similarity, and the intersection over union loss, of which combination is denoted as $\ell_{base}$. The total loss of the network is defined as:
 \begin{equation}
  \begin{aligned}
        \ell_{total}=\sum_{i=1}^{4} \frac{1}{2^{i}} \ell_{base}\left(S_{i}, G_i\right)+\ell_{base}\left(S_{out}, G\right),
  \end{aligned}
\end{equation}
 \begin{equation}
       \ell_{base}\left(S, G\right)=\ell_{bce}\left(S, G\right)+\ell_{ssim}\left(S, G\right)+\ell_{iou}\left(S, G\right),
\end{equation}
where $G$ denotes the corresponding ground truth, and $G_i$ is the side-output supervision, which is obtained by downsampling $G$ to a suitable size. Note that the loss functions of the side outputs set smaller weights to guide the training process.

\begin{table*}[!t]
	\caption{Quantitative comparison results in terms of S-measure ($S_\alpha$), max F-measure ($F_\beta$) and MAE score on five benchmark datasets. $\uparrow$ \& $\downarrow$ denote higher and lower is better, respectively. Bold number on each line represents the best performance.}
	\begin{center}
				 \renewcommand\arraystretch{1.1}
		
		\resizebox{\textwidth}{!}{
			\begin{tabular}{c|c|ccc|ccc|ccc|ccc|ccc}
		
			\toprule
\multicolumn{1}{c|}{\multirow{2}{*}{Method}} &\multicolumn{1}{c|}{\multirow{2}{*}{Pub‘Year}} & \multicolumn{3}{c|}{\textbf{DUT-test}}                                                                    & \multicolumn{3}{c|}{\textbf{LFSD}}                                                             &\multicolumn{3}{c|}{\textbf{NJU2K-test}}                                                             & \multicolumn{3}{c|}{\textbf{NLPR-test} }                                                                
& \multicolumn{3}{c}{\textbf{STERE1000}}                                                                  \\ \cline{3-17} 
\multicolumn{1}{c|}{}                        &\multicolumn{1}{c|}{}                        &
\multicolumn{1}{c}{$MAE\downarrow$} &
\multicolumn{1}{c}{$F_{\beta}\uparrow$} & \multicolumn{1}{c|}{$S_{\alpha}\uparrow$} & 

\multicolumn{1}{c}{$MAE\downarrow$} &
\multicolumn{1}{c}{$F_{\beta}\uparrow$} & \multicolumn{1}{c|}{$S_{\alpha}\uparrow$} & 

\multicolumn{1}{c}{$MAE\downarrow$} &
\multicolumn{1}{c}{$F_{\beta}\uparrow$} & \multicolumn{1}{c|}{$S_{\alpha}\uparrow$} & 

\multicolumn{1}{c}{$MAE\downarrow$} &
\multicolumn{1}{c}{$F_{\beta}\uparrow$} & \multicolumn{1}{c|}{$S_{\alpha}\uparrow$} & 
\multicolumn{1}{c}{$MAE\downarrow$}&
\multicolumn{1}{c}{$F_{\beta}\uparrow$} & \multicolumn{1}{c}{$S_{\alpha}\uparrow$}   \\
\hline
				
			DSA$^2$F~\cite{DSA2F}&CVPR'21
				&0.030 	&0.930 	&0.922 	
    &0.055 	&0.889 	&0.883 	
    &0.040 	&0.907 	&0.904 	
    &0.024 	&0.906 	&0.919 	
    &0.036 	&0.907 	&0.905 
 \\


			    DCF~\cite{DCF}&CVPR'21
			    &0.029 	&0.933 	&0.928 	
       &0.072 	&0.859 	&0.853 	
       &0.039 	&0.907 	&0.904 	
       &0.024 	&0.912 	&0.924 	
       &0.036 	&0.907 	&0.908 
 \\


				DFM-Net~\cite{DFMNet}&MM'21
			    &0.037 	&0.916 	&0.913 	
       &0.072 	&0.864 	&0.865 	
       &0.043 	&0.913 	&0.907 	
       &0.026 	&0.905   &0.923 	
       &0.045 	&0.893 	&0.898 \\

       BTS-Net~\cite{BTSNet}&ICME'21
				&0.048 	&0.889 	&0.894 	
    &0.071 	&0.868 	&0.865 	
    &0.035 	&0.927 	&0.925 	
    &0.023 	&0.917 	&0.931 	
    &0.038 	&0.910 	&0.913 
 \\

			TriTransNet~\cite{TriTransNet} &MM'21
				&0.025 	&0.944 	&0.933 	
    &0.066 	&0.870 	&0.866 	
    &0.030 	&0.926 	&0.919 	
    &0.021 	&0.921 	&0.929 	
    &0.033 	&0.911 	&0.908  \\

    VST~\cite{vst} &ICCV'21
				&0.024 	&0.947 	&\textbf{0.943 }	
    &0.054 	&0.892 	&\textbf{0.890 	}
    &0.035 	&0.919 	&0.922 	
    &0.023 	&0.917 	&0.932 	
    &0.038 	&0.907 	&0.913

 \\
    SP-Net~\cite{SPNet} &ICCV'21
				&0.047 	&0.894 	&0.890 	
    &0.068 	&0.867 	&0.860 	
    &\textbf{0.029} 	&0.928 	&0.925 	
    &0.022 	&0.914 	&0.926 	
    &0.037 	&0.906 	&0.907  \\

				CDNet~\cite{CDNet}&TIP'21
				&0.029 	&0.934 	&0.930 	
    &0.061 	&0.879 	&0.877 	
    &0.036 	&0.918 	&0.918 	
    &0.023 	&0.919 	&0.929 	
    &0.038 	&0.907 	&0.909 
 \\

				HAINet~\cite{HAINet}&TIP'21
				&0.034 	&0.927 	&0.919
    &0.072 	&0.862 	&0.859 	
    &0.038 	&0.910 	&0.910 	
    &0.025 	&0.905 	&0.921 	
    &0.038 	&0.909 	&0.909 
\\
    SPSN~\cite{SPSN} &ECCV'22
				&- 	&- 	&-
    &- 	&- 	&- 	
    &0.032 	&0.920 	&0.918 	
    &0.023 	&0.910 	&0.923 	
    &0.035 	&0.900 	&0.907 
  \\

    MVSalNet~\cite{MVSalNet} &ECCV'22
				&0.034 	&0.924 	&0.916 	
    &0.073 	&0.861 	&0.858 	
    &0.036 	&0.914 	&0.912 	
    &0.022 	&0.921 	&0.930 	
    &0.036 	&0.911 	&0.913  \\

				CCAFNet~\cite{CCAFNet}&TMM'22
				&0.036 	&0.915 	&0.905 	
       &0.087 	&0.832 	&0.827 	
       &0.037 	&0.911 	&0.910 	
       &0.026 	&0.909 	&0.922 	
       &0.044 	&0.887 	&0.891  \\
       
 RD3D~\cite{RD3D}&TNNLS'22
			    &0.029 	&0.936 	&0.931 	
       &0.074 	&0.854 	&0.858 	
       &0.037 	&0.914 	&0.915 	
       &0.022 	&0.916 	&0.930 	
       &0.037 	&0.906 	&0.911

  \\
 
CIRNet~\cite{crm/tip22/CIRNet} &TIP'22
				&0.029 	&0.938 	&0.932 	
    &0.068 	&0.883 	&0.875 	
    &0.035 	&0.928 	&0.925 	
    &0.022 	&0.921 	&0.933 	
    &0.039 	&0.913 	&0.916  \\

    DCMF~\cite{DCMF} &TIP'22
				&0.034 	&0.930 	&0.928 	
    &0.069 	&0.874 	&0.877 	
    &0.046 	&0.910 	&0.909 	
    &0.029 	&0.903 	&0.922 	
    &0.043 	&0.906 	&0.910  
    
     \\
       JL-DCF~\cite{JLDCF}&TPAMI'22
				&0.039 	&0.916 	&0.913 	
    &0.071 	&0.862 	&0.863 	
    &0.040 	&0.913 	&0.911 	
    &0.023 	&0.917 	&0.926 	
    &0.039 	&0.907 	&0.911 \\
    
	\hline		
OURS &--
                &\textbf{0.020} 	&\textbf{0.951} 	&\textbf{0.943 	}
                &\textbf{0.053} 	&\textbf{0.894} 	&0.888	
                &\textbf{0.029} 	&\textbf{0.931} 	&\textbf{0.927 	}
                &\textbf{0.019 }	&\textbf{0.928} 	&\textbf{0.935 	}
                &\textbf{0.031} 	&\textbf{0.920} 	&\textbf{0.921 }
 \\

				
				\bottomrule
			\end{tabular}}

	\end{center}
	\label{t1}
\end{table*}

\section{Experiments}
\subsection{Datasets and Evaluation metrics}

\begin{figure*}[!t]
\centering
\centerline{\includegraphics[width=1\textwidth]{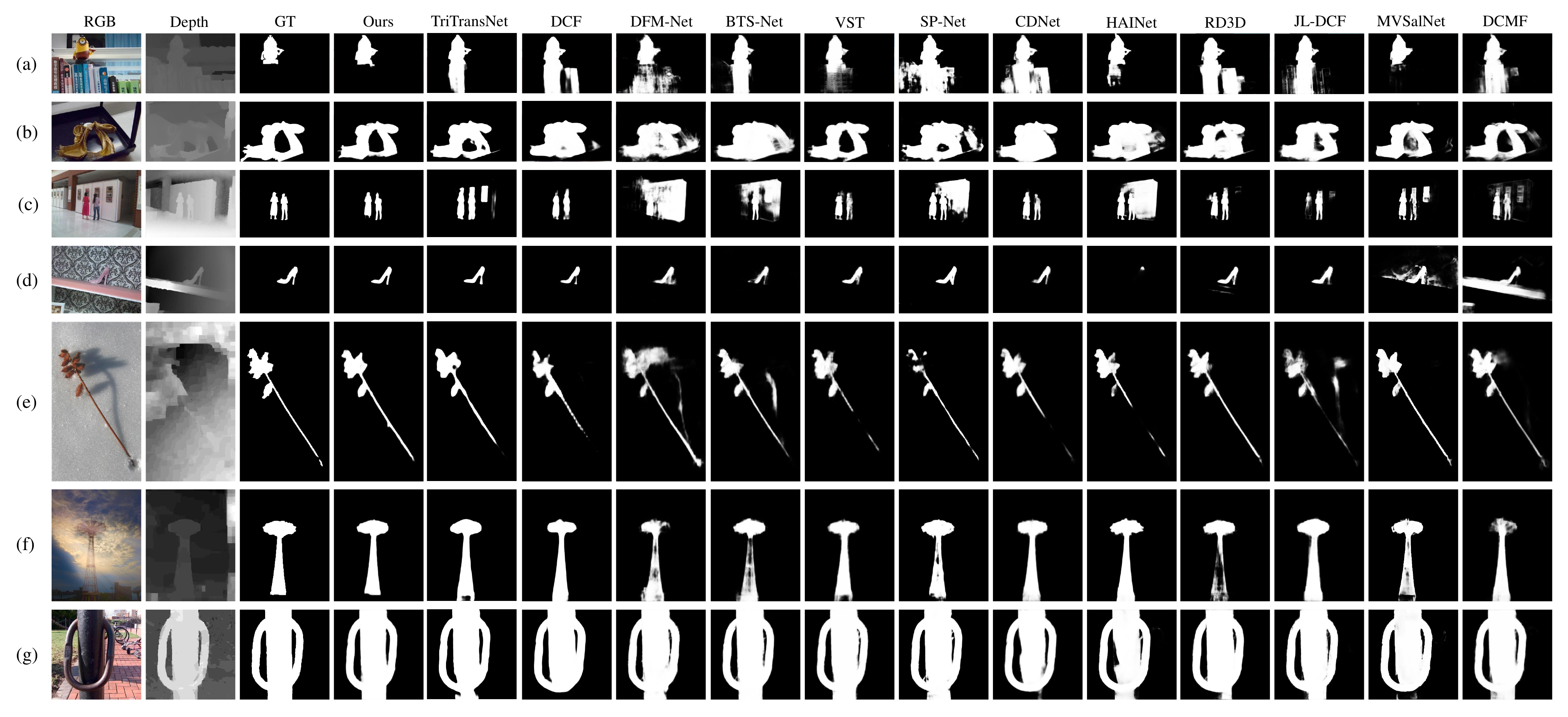}}
\caption{ Visual comparisons between our PICR-Net and SOTA methods under different challenging scenes, such as small objects (\ie, a, c and d), multiple objects (\ie, c), low contrast (\ie, d and f), low-quality depth map (\ie, b and e), and uneven lighting (\ie, g).}
\label{fig_result}
\vspace{0.2cm}
\end{figure*}

Five widely used RGB-D SOD benchmark datasets are employed to evaluate the performance of our PCIR-Net. The NLPR dataset~\cite{Peng2014} is obtained by Kinect 
camera, which contains $1000$ pairs of RGB images and depth maps from indoor and outdoor sences. 
Following~\cite{A2dele,cmMS}, we adopt $2985$ image pairs as our training data, including $1485$ samples from NJU2K dataset, $700$ samples from NLPR dataset, and $800$ samples from DUT dataset. All the remaining images in these training datasets, as well as LFSD~\cite{LFSD} and STERE1000~\cite{Niu2012} datasets are used for testing. 

We adopt three commonly used metrics in SOD task to quantitatively evaluate the performance. 
F-measure~\cite{Niu2012} indicates the weighted harmonic average of precision and recall by comparing the binary saliency map with ground truth.
MAE score~\cite{crm/tcsvt19/review} calculates the difference pixel by pixel.
S-measure~\cite{fan2017structure} evaluates the object-aware ($ S_o $) and region-aware structural ($ S_r $) similarity between the predicted saliency map and ground truth.




\subsection{Implementation Details}
The proposed network is implemented by the Pytorch and the MindSpore Lite tool$\footnote{https://www.mindspore.cn/}$, and uses a single NVIDIA GeForce RTX 3090 GPU for acceleration. All 
 the training and testing samples are resized to the size of $224\times224$ specified by Swin-Transformer. In addition, all depth maps are normalized and duplicated into three channels to fit the input size. Random flips and rotations are also used for data augmentation. During training, the encoder is initialized by parameters pre-trained on ImageNet. The Adam algorithm is used to optimize the proposed network with the batch size of $32$. The initial learning rate is set to $10^{-4}$, and a stepwise decay strategy is adopted, and every $40$ epoch decays to one-fifth of the previous one. The entire training process contains 90 epochs.

\subsection{Comparisons with the State-of-the-arts}
To prove the effectiveness of our proposed PICR-Net, we compare with 16 state-of-the-art models, including DSA$^2$F~\cite{DSA2F}, DCF~\cite{DCF}, DFM-Net~\cite{DFMNet}, TriTransNet~\cite{TriTransNet}, BTS-Net~\cite{BTSNet}, VST~\cite{vst}, SP-Net~\cite{SPNet}, CDNet~\cite{CDNet}, HAINet~\cite{HAINet}, CCAFNet~\cite{CCAFNet}, RD3D~\cite{RD3D}, JL-DCF~\cite{JLDCF}, SPSN~\cite{SPSN}, MVSalNet~\cite{MVSalNet}, CIRNet~\cite{crm/tip22/CIRNet} and DCMF~\cite{DCMF}. Among these, VST~\cite{vst} is a pure Transformer architecture, TriTransNet~\cite{TriTransNet} is a Transformer-assisted CNNs architecture, and the rest are pure CNNs-based architectures. For a fair comparison, we utilize the saliency maps provided by the authors or obtained from official testing codes for evaluation.

\subsubsection{Quantitative evaluation}
 Table~\ref{t1} intuitively shows the quantitative results of the proposed PICR-Net on five widely used datasets, where the best performance is marked in bold. Our proposed method outperforms all comparison methods on these five datasets, except for the S-measure on the LFSD dataset. 
 For example, compared with the second best method, the percentage gains of MAE score reach 16.7\%, 1.9\%, 9.5\%, and 6.1\% on the DUT-test, LFSD, NLPR-test, and STERE1000 datasets, respectively. Similar gains can be observed in other metrics.
 \begin{table}[!t]
	\scriptsize
	\caption{Inference speed of our PICR-Net and some typical SOTA methods. Black bold fonts indicate the best performance }
	\begin{center}
				 \renewcommand\arraystretch{1.1}
		\setlength{\tabcolsep}{1.5mm}{
		\vspace{0.2cm}
			\begin{tabular}{c|c|ccc|ccc}
			\toprule
\multicolumn{1}{c|}{\multirow{2}{*}{Method}}& \multirow{2}{*}{Speed (FPS)} & \multicolumn{3}{c|}{\textbf{NJU2K-test}}  & \multicolumn{3}{c}{\textbf{NLPR-test}}                                                                                                                                                                 \\ \cline{3-8} 

\multicolumn{1}{c|}{}        &        \multicolumn{1}{c|}{}                              &
\multicolumn{1}{c}{${MAE}\downarrow$} &
\multicolumn{1}{c}{$F_{\beta}\uparrow$} & 
\multicolumn{1}{c|}{$S_{\alpha}\uparrow$} &
\multicolumn{1}{c}{${MAE}\downarrow$} &
\multicolumn{1}{c}{$F_{\beta}\uparrow$} & 
\multicolumn{1}{c}{$S_{\alpha}\uparrow$}   \\ \toprule
               
TriTransNet & 16.38

    &0.030 	&0.926 	&0.919 	
    &0.021 	&0.921 	&0.929 	
 \\

    VST & 15.74

    &0.035 	&0.919 	&0.922 	
    &0.023 	&0.917 	&0.932 	 \\

SP-Net & 13.10
				
    &\textbf{0.029} 	&0.928 	&0.925 	
    &0.022 	&0.914 	&0.926 	
 \\

               \hline
                 Ours& \textbf{21.29}
                &\textbf{0.029}	&\textbf{0.931}	&\textbf{0.927}
                &\textbf{0.019 }	&\textbf{0.928} 	&\textbf{0.935} \\
				\bottomrule
			\end{tabular}}

	\end{center}
	\label{speed}
\end{table}
 Inference speed has always been a key factor restricting the development and application of deep learning models\cite{caaitrit/WuZLW22}. So we also evaluate the inference speed of our PICR-Net and other typical SOTA models including Transformer-based model VST~\cite{vst}, TriTransNet~\cite{TriTransNet} and advanced CNNs-based model SP-Net \cite{SPNet}. As shown in Table~\ref{speed}, our model achieves better performance while also having an advantage in inference speed. However, our model has not yet achieved real-time efficiency, which is also a research point for further improving the inference speed of the Transformer-based model in the future.

\subsubsection{Qualitative comparison}
Figure~\ref{fig_result} provides some visualization results of different methods, including challenging scenarios with small objects (\ie, a, c and d), multiple objects (\ie, c), low contrast (\ie, d and f), low-quality depth map (\ie, b and e), and uneven lighting (\ie, g). As can be seen, our method not only accurately detects salient objects in those challenging scenarios, but also obtains better completeness and local details. 
It is worth noting that Transformer-based models (\ie, VST, TriTransNet, and our PICR-Net) are able to model global dependencies and therefore tend to outperform the rest of the CNNs-based networks in terms of salient object localization. 
In addition, thanks to the well-designed cross-modality interaction, our network can fully extract information from the other modality to achieve accurate and complete prediction when the quality of the depth map is relatively poor (\eg, Figure~\ref{fig_result}(a) and (e)) or there is light and shadow interference in RGB image (\eg, Figure~\ref{fig_result}(g)). 
At the same time, because the CNNR unit provides more fine-grained detail information, compared with other methods, our method has more advantages in boundary accuracy and detail description (\eg, Figure~\ref{fig_result}(c), (d) and (g)).
Both quantitative and qualitative experiments above demonstrate the effectiveness of our proposed method.

\subsection{Ablation Studies}
We conduct ablation experiments on the NJU2K-test and NLPR-test dataset to verify the role of each module in the proposed PICR-Net.

\subsubsection{Effectiveness of general structure}
\begin{figure}[!t]
  \centering
  \includegraphics[width=0.95\linewidth]{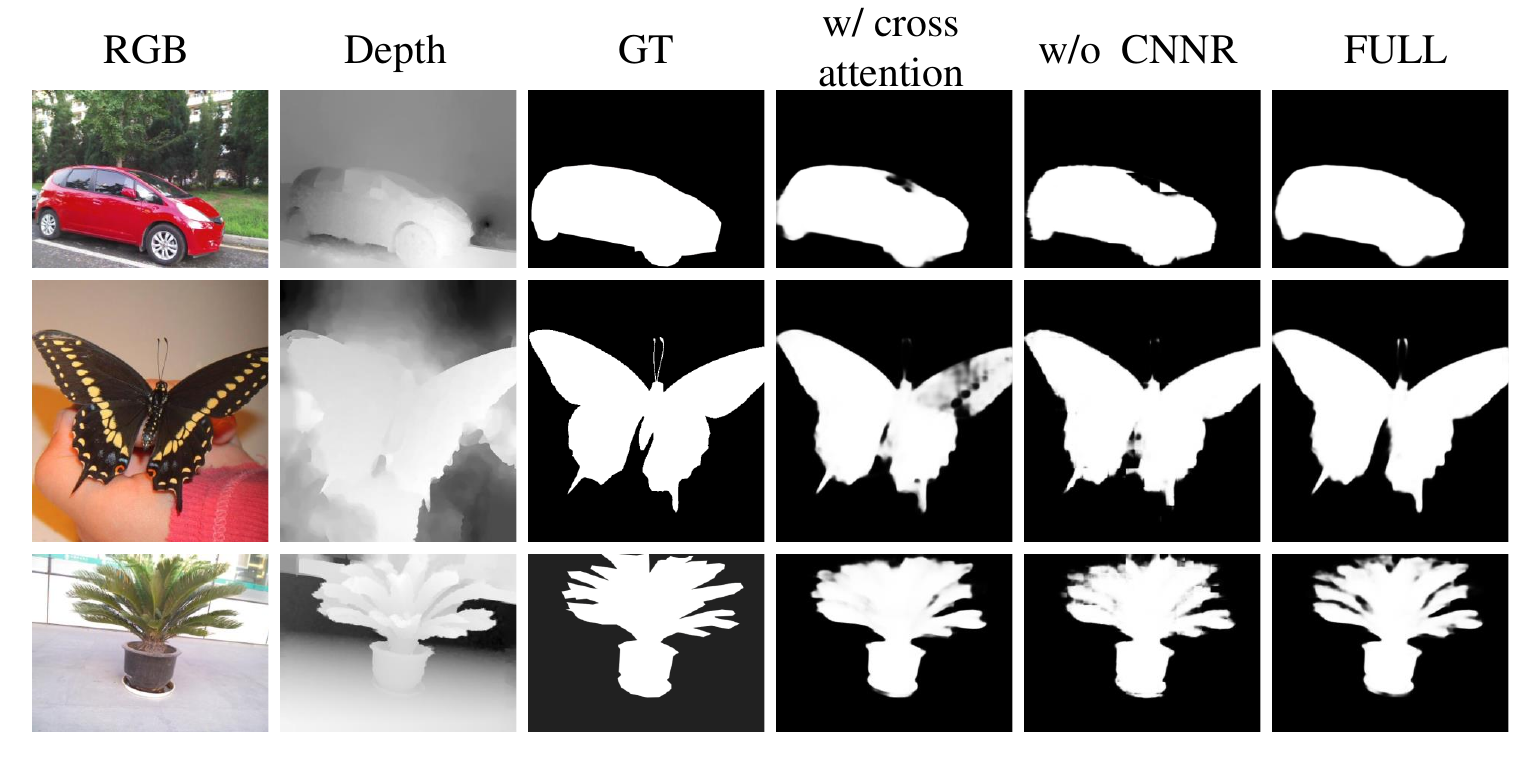}
  \caption{Visual comparisons of different ablation studies.}
  \label{abla_cru}
\end{figure}

\begin{table}[!t]
	\scriptsize
	\caption{Quantitative ablation evaluation of general structure. Black bold fonts indicate the best performance. }
	\begin{center}
				 \renewcommand\arraystretch{1.1}
		\setlength{\tabcolsep}{1.5mm}{
		\vspace{0.2cm}
			\begin{tabular}{c|c|ccc|ccc}
			\toprule
\multicolumn{1}{c|}{\multirow{2}{*}{Method}}& \multirow{2}{*}{ID} & \multicolumn{3}{c|}{\textbf{NJU2K-test}}  & \multicolumn{3}{c}{\textbf{NLPR-test}}                                                                                                                                                                 \\ \cline{3-8} 

\multicolumn{1}{c|}{}        &        \multicolumn{1}{c|}{}                              &
\multicolumn{1}{c}{${MAE}\downarrow$} &
\multicolumn{1}{c}{$F_{\beta}\uparrow$} & 
\multicolumn{1}{c|}{$S_{\alpha}\uparrow$} &
\multicolumn{1}{c}{${MAE}\downarrow$} &
\multicolumn{1}{c}{$F_{\beta}\uparrow$} & 
\multicolumn{1}{c}{$S_{\alpha}\uparrow$}   \\ \toprule
                FULL&0 
                &\textbf{0.029}	&\textbf{0.931}	&\textbf{0.927}
                &\textbf{0.019 }	&\textbf{0.928} 	&\textbf{0.935}

               \\\hline

				w/ addition & 1
              &0.038 	&0.907 	&0.909 
 	  	&0.023 	&0.914 	&0.925 
 	  	\\
                w/ multiplication & 2
             &0.035 	&0.918 	&0.916 
            &0.022	&0.918 	&0.927

 	  	\\
     w/ concatenation & 3
             &0.033 &0.921 	&0.918 	
             &0.024  &0.912 	&0.925

 	  	\\
      w/ cross-attention & 4
             &0.031 	&0.925 	&0.922 	
             &\textbf{0.019} 	&0.924 	&0.933

 	  	\\
     \hline

				w/o TD & 5
              &0.034 	&0.921 	&0.919 &0.022 	&0.916 	&0.928 

 	  	\\
      w/o CNNR & 6
             &0.031	&0.925 	&0.923 	&0.020 	&0.923 	&0.933 
	 	\\
				\bottomrule
			\end{tabular}}

	\end{center}
	\label{abla1}
\end{table}
First, in order to verify the role of the CmPI module, we design the following substitution experiments:
\begin{itemize}
\item FULL (id 0) means our proposed full model PICR-Net.
  \item w/ addition (id 1), w/ multiplication (id 2) and w/ concatenation (id 3) respectively indicate that the CmPI module is replaced by element-level addition, multiplication, and concatenation operations to achieve interaction between RGB and depth features.
  \item w/ cross-attention (id 4) means using traditional cross-attention~\cite{vst} operation to replace RM in the CmPI module.
  
\end{itemize} 
As shown in Table~\ref{abla1}, our designed CmPI module achieves better performance than other simple interaction strategies. Also, comparing id 0 and id 4, it can be found that our full model with CmPI module achieves better performance, and also outperforms cross-attention which brings more computational burden.
Figure~\ref{abla_cru} provides some visualization results of different ablation studies. From the second image, it can be seen that low-quality depth map can negatively impact interactions using cross-attention, leading to object omission. In contrast, our method still detects salient objects accurately and completely.

Besides, in order to verify the effectiveness of the Transformer-based decoder with CNNR unit, we design the following stripping experiments:
\begin{itemize}
  \item w/o TD (id 5) replaces Transformer-based decoder with equal number of convolution layers for saliency decoding.
  \item w/o CNNR (id 6) removes the CNNR unit at the end of the decoder.
\end{itemize} 
As in Table~\ref{abla1}, after the replacement for Transformer-based decoder, the F-measure scores on two datasets are decreased by $ 1.1\% $ and $ 1.3\% $, respectively, demonstrating that using CNNs to complete decoding will dilute the global information extracted by the Transformer and reduce the performance. 
In addition, it can be found in Figure~\ref{abla_cru} that CNNR unit contributes to the improvement of the boundary quality and the clarity of the saliency maps, which is also supported by the quantitative results. 

\subsubsection{Effectiveness of design detail}
Moreover, to verify the effectiveness of the detailed design of the CmPI module, we design the following experiments:
\begin{itemize}
  \item w/o RM (id 7) removes the key component RM in the CmPI module.
  \item w/ single-step (id 8) means only keeping the first attention operation in RM, that is to say, the second-step attention calculation is removed.
  \item w/o $M_1\& M_2$ (id 9) removes the mask constraint in the calculation Eq. (6) of RM, that is, the $M_1$ and $M_2$ are removed. In addition, w/o $M_1$ (id 10) and w/o $M_2$ (id 11) represent the removal of only $M_1$ and $M_2$ respectively.
  \item w/o $g_{r/d}$ (id 12) removes the global guidance vector $g_{r}$ and $g_{d}$ in RM.
  \item Win\_3 (id 13) and Win\_5 (id 14) mean that the window size for attention interaction in RM is adjusted from 1(point-aware) to 3 and 5, respectively.
  \item $1\times1$ convolution  (id 15) replaces RM with $1\times1$ convolution which is also point-aware operation.
\end{itemize}
\begin{table}[!t]
	\scriptsize
	\caption{Quantitative ablation evaluation of detailed design in CmPI. Black bold fonts indicate the best performance }
	\begin{center}
				 \renewcommand\arraystretch{1.1}
		\setlength{\tabcolsep}{1.5mm}{
		\vspace{0.2cm}
			\begin{tabular}{c|c|ccc|ccc}
			\toprule
\multicolumn{1}{c|}{\multirow{2}{*}{Method}}& \multirow{2}{*}{ID} & \multicolumn{3}{c|}{\textbf{NJU2K-test}}  & \multicolumn{3}{c}{\textbf{NLPR-test}}                                                                                                                                                                 \\ \cline{3-8} 

\multicolumn{1}{c|}{}        &        \multicolumn{1}{c|}{}                              &
\multicolumn{1}{c}{${MAE}\downarrow$} &
\multicolumn{1}{c}{$F_{\beta}\uparrow$} & 
\multicolumn{1}{c|}{$S_{\alpha}\uparrow$} &
\multicolumn{1}{c}{${MAE}\downarrow$} &
\multicolumn{1}{c}{$F_{\beta}\uparrow$} & 
\multicolumn{1}{c}{$S_{\alpha}\uparrow$}   \\ \toprule
                FULL&0 
                &\textbf{0.029}	&\textbf{0.931}	&\textbf{0.927}
                &\textbf{0.019 }	&\textbf{0.928} 	&\textbf{0.935}

               \\\hline

       w/o RM  &7
             &0.035 	&0.916 	&0.916 
 	 	&0.023 	&0.919 	&0.928 
	 	\\
   w/ single-step  & 8
             &0.030 	&0.930 	&0.926 
 	 	&0.020 	&0.925 	&0.932 
	 	\\
    w/o $M_1\& M_2$ & 9
             &0.032	&0.923 	&0.921 
 	 	&0.021 	&0.921 	&0.929 
	 	\\
   
                w/o $M_1$  & 10
             &0.031 	&0.923 	&0.923 	 &0.020 	&0.924 	&0.933

 	  	\\
       w/o $M_2$  & 11
        &0.030 	&0.927 	&0.925 &0.020	&0.923 	&0.932
             \\
    w/o $g_{r/d}$  & 12
               &0.030 	&0.926 	&0.924 	&\textbf{0.019} 	&0.924 	&0.934

 	  	\\
          Win\_3  & 13
    &\textbf{0.029} 	&0.930 	&\textbf{0.927 }	  &0.020 	&0.926 	&0.934

 	  	\\
  
      Win\_5  & 14
      &\textbf{0.029} 	&0.929 	&0.926 &0.020 	&0.924 	&0.932

 	  	\\
     $1\times1$ convolution  & 15
   &0.031 	&0.926 	&0.922  &0.020 	&0.924 	&0.932  \\

				\bottomrule
			\end{tabular}}

	\end{center}
	\label{abla}
\end{table}
\begin{figure}[!h]
  \centering
  \includegraphics[width=1\linewidth]{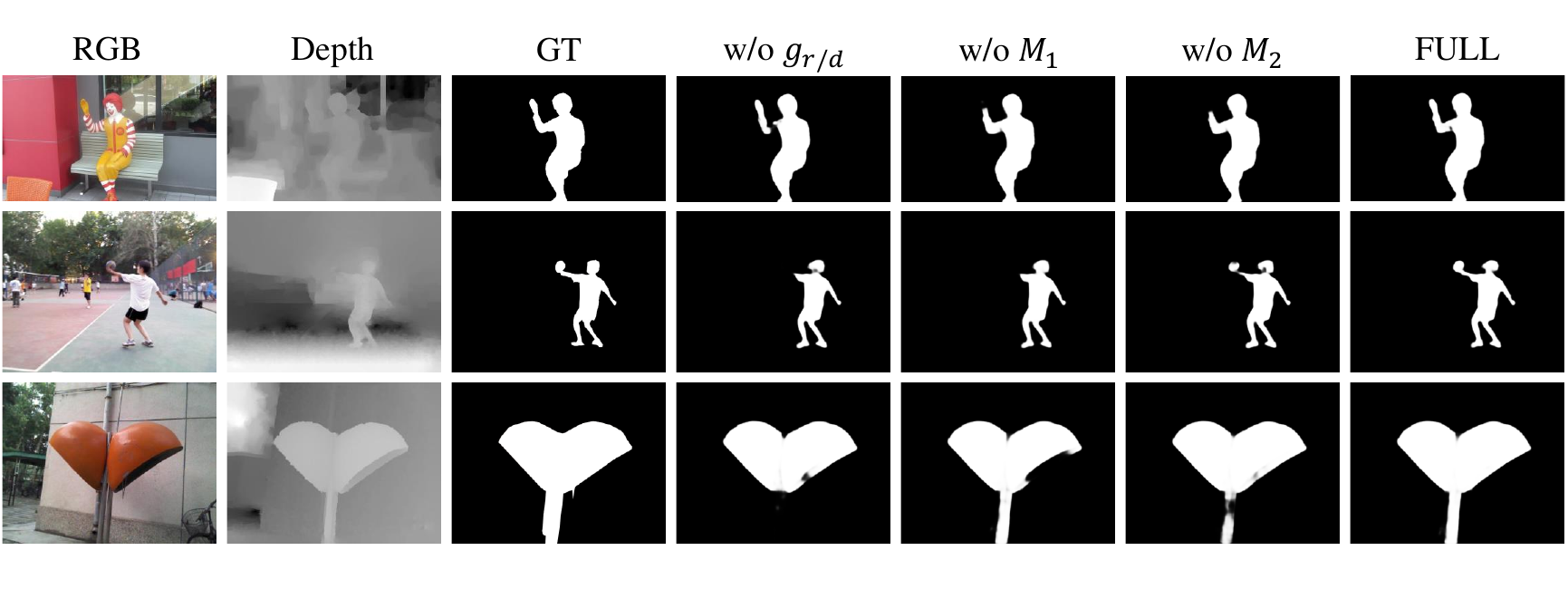}

  \caption{Qualitative comparisons of ablation studies on design detail in CmPI.}
  \label{abla_detail}
\end{figure}
The related results are reported in Table \ref{abla}.
Overall, all ablation validations are inferior to our FULL model design. Specifically, if the entire RM module is removed, the performance loss is very obvious, as show in the result of id 7. In addition, the self-modality global-local guidance in the second-step attention (as shown in the result of id 8) and the suppression of negative interactions in RM (as shown in the result of id 9, 10 and 11) are both very necessary and effective. For the guidance vectors $g_{r}$ and $g_{d}$, after removing them, the performance drops, which also leads to worse object integrity as shown in Figure~\ref{abla_detail}. For the interaction range of attention operations, enlarging the window size (3 or 5) does not obviously improve performance due to the strong correlation of positions, but brings exponential computational cost. We directly use $1\times 1$ convolution to replace the CmPI module (for a fair comparison, guidance vectors are also introduced by expansion and concatenation), as shown in id 15 of Table~\ref{abla}. All metrics are reduced on both datasets, indicating that CmPI can achieve more comprehensive interaction than $1\times 1$ convolution.

\section{Conclusion}

Considering the respective characteristics and advantages of Transformer and CNNs, we propose a network named PICR-Net to achieve RGB-D SOD, where the network follows encoder-decoder architecture based on Transformer as a whole, and adds a pluggable CNNR unit at the end for detail refinement.
Moreover, compared with the traditional cross-attention, our proposed CmPI module considers the prior correlation between RGB and depth modalities, enabling more effective cross-modality interaction by introducing spatial constraints and global saliency guidance.
The comprehensive experiments demonstrate that our network achieves competitive performance against 16 state-of-the-art methods on five benchmark datasets.






\begin{acks}
This work was supported in part by National Natural Science Foundation of China under Grant 61991411, in part by the Taishan Scholar Project of Shandong Province under Grant tsqn202306079, in part by Project for Self-Developed Innovation Team of Jinan City under Grant 2021GXRC038, in part by the National Natural Science Foundation of China under Grant 62002014, in part by the Hong Kong Innovation and Technology Commission (InnoHK Project CIMDA), in part by the Hong Kong GRF-RGC General Research Fund under Grant 11203820 (CityU 9042598), in part by Young Elite Scientist Sponsorship Program by the China Association for Science and Technology under Grant 2020QNRC001, and in part by CAAI-Huawei MindSpore Open Fund.
\end{acks}

\bibliographystyle{ACM-Reference-Format}
\balance
\bibliography{ref}

\newpage
\appendix
\section{Appendix}
\subsection{Hyperparameter setting}
The -100 is a hyperparameter in attention mask $M_1$ and $M2$, which aims to suppress the irrelevant negative interactions in RM. And it is presented exponentially in the subsequent softmax operation, in this way, -100 is small enough to suppress those negative interactions. We also perform experiments to verify the effect of this hyperparameter by replacing -100 with -10, -1000, and -10000, respectively. As shown in Table~\ref{hyper}, the -100 will perform better than the -10 but further tuning it down will not noticeably improve performance. This also indirectly explains the rationality of our parameter settings.
\begin{table}[h]
	\scriptsize
	\caption{Quantitative comparison on hyperparameters in attention masks}
	\begin{center}
				 \renewcommand\arraystretch{1.1}
		\setlength{\tabcolsep}{1.5mm}{
		\vspace{0.2cm}
			\begin{tabular}{c|ccc|ccc}
			\toprule
\multicolumn{1}{c|}{\multirow{2}{*}{Method}} & \multicolumn{3}{c|}{\textbf{NJU2K-test}}  & \multicolumn{3}{c}{\textbf{NLPR-test}}                                                                                                                                                                 \\ \cline{2-7} 

\multicolumn{1}{c|}{}                                     &
\multicolumn{1}{c}{${MAE}\downarrow$} &
\multicolumn{1}{c}{$F_{\beta}\uparrow$} & 
\multicolumn{1}{c|}{$S_{\alpha}\uparrow$} &
\multicolumn{1}{c}{${MAE}\downarrow$} &
\multicolumn{1}{c}{$F_{\beta}\uparrow$} & 
\multicolumn{1}{c}{$S_{\alpha}\uparrow$}   \\ \toprule
               
-10 

    &0.029 	&0.930 	&0.927 	
    &0.020 	&0.925 	&0.934 	
 \\

    -100(PICR-Net)

    &0.029 	&0.931 	&0.927 	
    &0.019 	&0.926 	&0.934 	 	 \\

-1000
				
    &0.030 	&0.930 	&0.924	
    &0.018 	&0.930 	&0.937 	
 \\

                 -10000
                &0.029 	&0.930 	&0.926	
    &0.019 	&0.926 	&0.934 	 \\
				\bottomrule
			\end{tabular}}

	\end{center}
	\label{hyper}
\end{table}

\subsection{Fusion stage}
 In our PICR-Net, we perform cross-modal feature interaction in the decoding stage. To verify the reasonableness of doing so, we attempt to implement cross-modality interactions between the encoder layers, and use skip connections without any interactive operations in the decoding stage.
 As shown in Table~\ref{stage}, cross-modal fusion in the encoding stage is not as effective as in the decoding stage. Specifically, ${MAE}$, {$F_{\beta}$}, and $S_{\alpha}$ decreased by 10.52\%, 0.96\%, and 0.64\%, respectively, on the NLPR-test dataset. The reason for this is that fusion in the encoding stage taking in shallow-deep way, and the presence of a large amount of noise in the shallow features can cause interference.

\begin{table}[h]
	\scriptsize
	\caption{Quantitative comparison on different fusion stage}
	\begin{center}
				 \renewcommand\arraystretch{1.1}
		\setlength{\tabcolsep}{1.5mm}{
		\vspace{0.2cm}
			\begin{tabular}{c|ccc|ccc}
			\toprule
\multicolumn{1}{c|}{\multirow{2}{*}{Fusion Stage}} & \multicolumn{3}{c|}{\textbf{NJU2K-test}}  & \multicolumn{3}{c}{\textbf{NLPR-test}}                                                                                                                                                                 \\ \cline{2-7} 

\multicolumn{1}{c|}{}                                     &
\multicolumn{1}{c}{${MAE}\downarrow$} &
\multicolumn{1}{c}{$F_{\beta}\uparrow$} & 
\multicolumn{1}{c|}{$S_{\alpha}\uparrow$} &
\multicolumn{1}{c}{${MAE}\downarrow$} &
\multicolumn{1}{c}{$F_{\beta}\uparrow$} & 
\multicolumn{1}{c}{$S_{\alpha}\uparrow$}   \\ \toprule
               
Encoder 

    &0.034 	&0.925 	&0.919 	
    &0.021 	&0.919 	&0.929 	
 \\

    Decoder(PICR-Net)

    &0.029 	&0.931 	&0.927 	
    &0.019 	&0.926 	&0.934 	 	 \\

				\bottomrule
			\end{tabular}}

	\end{center}
	\label{stage}
\end{table}

\end{sloppypar}

\end{document}